\documentclass[10pt,twocolumn,letterpaper]{article}

\usepackage{cvpr}
\usepackage{times}
\usepackage{epsfig}
\usepackage{graphicx}
\usepackage{amsmath}
\usepackage{amssymb}
\usepackage{caption}
\usepackage{latexsym}
\usepackage{amsfonts}
\usepackage{xcolor}
\usepackage{soul}
\usepackage{appendix}
\usepackage{tabularx}
\usepackage{anyfontsize}

\usepackage[belowskip=0pt,aboveskip=0pt,font=small]{caption}
\usepackage[belowskip=0pt,aboveskip=0pt,font=small]{subcaption}
\usepackage{float}
\usepackage{multirow}
\usepackage{rotating}
\usepackage{booktabs}
\usepackage{mysymbols}

\usepackage[pagebackref=true,breaklinks=true,letterpaper=true,colorlinks,bookmarks=false]{hyperref}

\cvprfinalcopy 

\def\cvprPaperID{****} 

\begin{document}

\title{Multi-Target Embodied Question Answering}

\author{Licheng Yu$^1$, Xinlei Chen$^3$, Georgia Gkioxari$^3$, Mohit Bansal$^1$, \\Tamara L. Berg$^{1,3}$,
Dhruv Batra$^{2,3}$\\ 
{$^1$University of North Carolina at Chapel Hill
~~~$^2$Georgia Tech
~~~$^3$Facebook AI}
}


\maketitle

\begin{abstract}
Embodied Question Answering (EQA) is a relatively new task where an agent is asked to answer questions about its environment from egocentric perception. 
EQA as introduced in \cite{das2018embodied} makes the fundamental assumption that every question, \eg ``what color is the car?", has exactly \textbf{one} target (``car") being inquired about. This assumption puts a direct limitation on the abilities of the agent.

We present a generalization of EQA -- Multi-Target EQA (MT-EQA). 
Specifically, we study questions that have \textbf{multiple} targets in them, such as ``Is the dresser in the bedroom bigger than the oven in the kitchen?", where the agent has to navigate to multiple locations (``dresser in bedroom", ``oven in kitchen") and  perform comparative reasoning (``dresser" bigger than ``oven") before it can answer a question. Such questions require the development of entirely new modules or components in the agent. 
To address this, we propose a modular architecture composed of a program generator, a controller, a navigator, and a VQA module. 
The program generator converts the given question into sequential executable sub-programs; the navigator guides the agent to multiple locations pertinent to the navigation-related sub-programs; and the controller learns to select relevant observations along its path. These observations are then fed to the VQA module to predict the answer. 
We perform detailed analysis for each of the model components and show that our joint model can outperform previous methods and strong baselines by a significant margin.
Project page: \href{https://embodiedqa.org}{https://embodiedqa.org}.
\end{abstract}

\vspace{-.5cm}
\section{Introduction}
\label{sec:Intro}
\vspace{-.1cm}


\begin{figure}[t]
\centering
\includegraphics[width=0.48\textwidth]{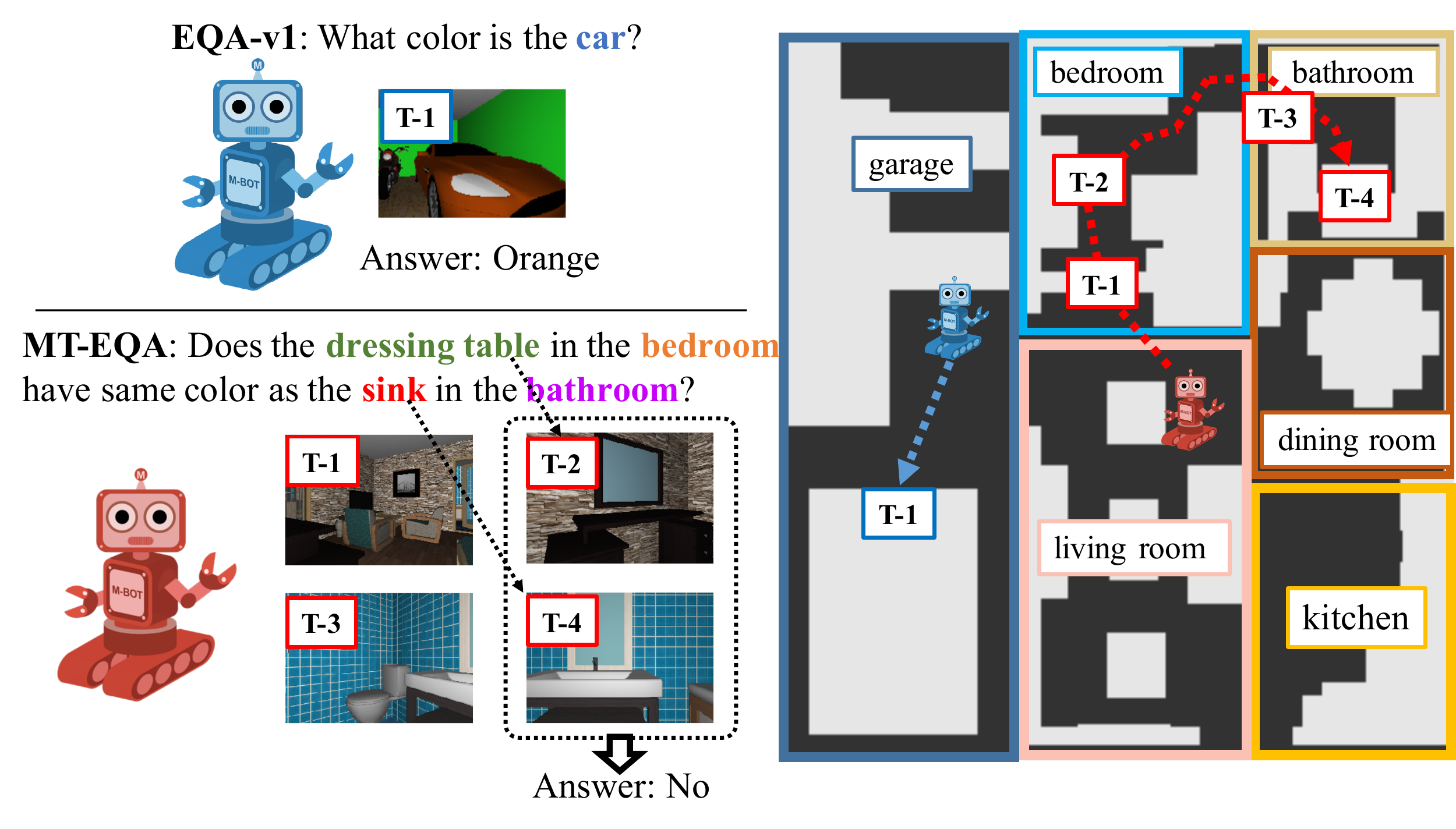}
\caption{Difference between EQA-v1 and MT-EQA. While EQA-v1's question asks about a single target ``car", MT-EQA's question involves multiple targets (e.g., bedroom, dressing table, bathroom, sink) to be navigated, and attribute comparison between multiple targets (e.g., dressing table and sink).}
\label{fig:highlight}
\vspace{-0.4cm}
\end{figure}

One of the grand challenges of AI is to build intelligent agents that visually perceive their surroundings, communicate with humans via natural language, and act in their environments to accomplish tasks.
In the vision, language, and AI communities, we are witnessing a shift in focus from \emph{internet vision} to \emph{embodied AI} -- with the creation of new tasks and benchmarks~\cite{chaplot2017gated, anderson2018vision, gupta2017cognitive, zhu2017visual}, instantiated on new simulation platforms~\cite{kolve2017ai2, savva2017minos, wu2018building, xia2018gibson, kempka2016vizdoom,brodeur2017home}.

The focus of this paper is one such embodied AI task, Embodied Question Answering (EQA)~\cite{das2018embodied}, which tests an agent's overall ability to jointly perceive its surrounding, communicate with humans, and act in a physical environment. 
Specifically, in EQA, an agent is spawned in a random location within an environment and is asked a question about something in that environment, for example \myquote{What color is the lamp?}. 
In order to answer the question correctly, the agent needs to parse and understand the question, navigate to a good location (looking at the ``lamp'') based on its first-person perception of the environment and predict the right answer (\eg ``blue'').

However, there is still much left to be done in EQA. 
In its original version, the EQA-v1 dataset only consists of single-target question-answer pairs, such as \myquote{What color is the car?}.
The agent just needs to find the car then check its color based on its last observed frames.
However, the single target constraint places a direct limitation on the possible set of tasks that the AI agent can tackle. For example, consider the question \myquote{Is the kitchen larger than the bedroom?} in EQA-v1; the agent would not be able to answer this question because it involves navigating to multiple targets --``kitchen'' \emph{and} ``bedroom'' -- and the answer requires comparative reasoning between the two rooms, where all of these skills are not part of the original EQA task.

In this work, we present a generalization of EQA -- multi-target EQA (MT-EQA). 
Specifically, we study questions that have multiple implicit targets in them, such as \myquote{Is the dresser in the bedroom bigger than the oven in the kitchen?}.
At a high-level, our work is inspired by the visual reasoning work of Neural Modular Networks~\cite{andreas2016neural} and CLEVR~\cite{johnson2017clevr}.
These works study compositional and modular reasoning in a fully-observable environment (an image).
Our work may be viewed as embodied visual reasoning, where an agent is asked a question involving multiple modules and needs to gather information before it can execute them.
In MT-EQA, we propose 6 types of compositional questions which compare attribute properties (color, size, distance) between multiple targets (objects/rooms).
Fig.~\ref{fig:highlight} shows an example from the MT-EQA dataset and contrasts it to the original EQA-v1 dataset.

\begin{figure}[t]
\centering
\includegraphics[width=0.45\textwidth]{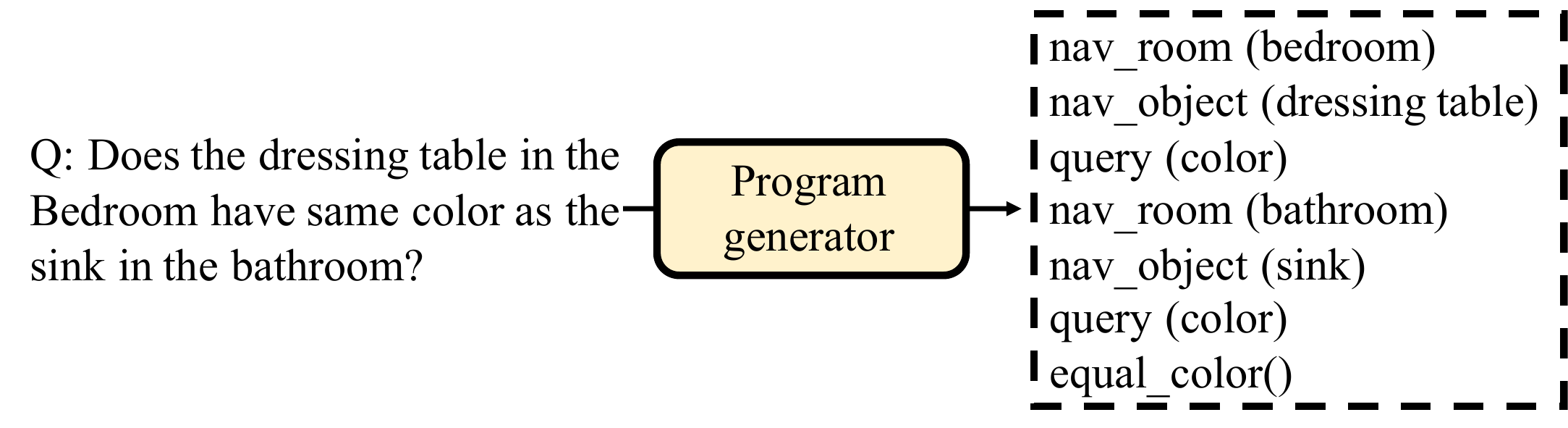}
\vspace{-0.0cm}
\caption{Program Generator.}
\vspace{-0.2cm}
\label{fig:pg}
\end{figure}

The assumption in EQA-v1 of decoupling navigation from question-answering not only makes the task simpler but is also reflected in the model used -- the EQA-v1 model simply consists of an LSTM navigator which after stopping, hands over frames to a VQA module. 
In contrast, MT-EQA introduces new modeling challenges that we address in this work. 
Consider the MT-EQA question in Fig.~\ref{fig:highlight} -- \myquote{Does the table in the bedroom have same color as the sink in the bathroom?}. From this example, it is clear that not only is it necessary to have a tighter integration between navigator and VQA, but we also need to develop fundamentally new modules.
An EQA-v1~\cite{das2018embodied} agent would navigate to the final target location and run the VQA module based on its last sequence of frames along the path. 
In this case, only the ``sink" would be observed from the final frames but dressing table would be lost.
Instead, we propose a new model that consists of 4 components: (a) a program generator, (b) a navigator, (c) a controller and (d) a VQA module. 
The program generator converts the given question into sequential executable sub-programs, as shown in Fig.~\ref{fig:pg}. 
The controller executes these sub-programs sequentially and gives control to the navigator when the navigation sub-programs are invoked (\eg \texttt{nav\_room(bedroom)}). 
During navigation, the controller processes the first-person views observed by the agent and predicts whether the target of the sub-program (\eg bedroom) has been reached. 
In addition, the controller extracts cues pertinent to the questioned property of the sub-target, \eg \texttt{query(color)}. 
Finally, these cues are fed into the VQA module which deals with the comparison of different attributes, \eg  executing \texttt{equal\_color()} by comparing the color of dressing table and sink (Fig.~\ref{fig:highlight} ).

Empirically, we show results for our joint model and analyze the performance of each of our components.
Our full model outperforms the baselines under almost every navigation and QA metric by a large margin.
We also report performance for the navigator, the controller, and the VQA module, when executed separately in an effort to isolate and better understand the effectiveness of these components.
Our ablation studies show that our full model is better at all sub-tasks, including room navigation, object navigation and final EQA accuracy.
Additionally, we find quantitative evidence that MT-EQA questions on closer targets are relatively easier to solve as they require shorter navigation, while questions for farther targets are harder.



\vspace{-.1cm}
\section{Related Work}
\label{sec:related}
\vspace{-.1cm}
Our work relates to research in embodied perception and modular predictive models for program execution.

\smallskip
\noindent {\bf Embodied Perception.}
Visual recognition from images has witnessed tremendous success in recent years with the advent of deep convolutional neural networks (CNNs)~\cite{krizhevsky2012imagenet, Szegedy2015, He2016} and large-scale datasets, such as ImageNet~\cite{Russakovsky2015} and COCO~\cite{Lin2014}. More recently, we are beginning to witness a resurgence of \emph{active vision}. For example, end-to-end learning methods successfully predict robotic actions from raw pixel data~\cite{Levine2016}. Gupta \etal~\cite{gupta2017cognitive} learn to navigate via mapping and planning. Sadeghi \& Levine~\cite{sadeghi2017cadrl} teach an agent to fly in simulation and show its performance in the real world. Gandhi \etal~\cite{gandhi2017} train self-supervised agents to fly from examples of drones crashing.

At the intersection of active perception and language understanding, several tasks have been proposed, including instruction-based navigation~\cite{chaplot2017gated,anderson2018vision}, target-driven navigation~\cite{zhu2017target,gupta2017cognitive}, embodied question answering~\cite{das2018embodied}, interactive question answering~\cite{gordon2018iqa}, and task planning~\cite{zhu2017visual}.
While these tasks are driven by different goals, they all require training agents that can perceive their surroundings, understand the goal -- either presented visually or in language instructions -- and act in a virtual environment. Furthermore, the agents need to show strong generalization ability when deployed in novel unseen environments~\cite{gupta2017cognitive, wu2018building}.


\smallskip
\noindent {\bf Environments.}
There is an overbearing cost to developing real-world interactive benchmarks. 
Undoubtedly, this cost has hindered progress in studying embodied tasks. 
On the contrary, virtual environments that offer rich, efficient simulations of real-world dynamics, have emerged as promising alternatives to potentially overcome many of the challenges faced in real-world settings.

Recently there has been an explosion of simulated 3D environments in the AI community, all tailored towards different skill sets. Examples include ViZDoom~\cite{kempka2016vizdoom}, TorchCraft~\cite{synnaeve2016torchcraft} and DeepMind Lab~\cite{beattie2016deepmind}. 
Just in the last year, simulated environments of semantically complex, realistic 3D scenes have been introduced, such as  HoME~\cite{brodeur2017home}, House3D~\cite{wu2018building}, MINOS~\cite{savva2017minos}, Gibson~\cite{xia2018gibson} and AI2THOR~\cite{kolve2017ai2}.
In this work, we use House3D, following the original EQA task~\cite{das2018embodied}.
House3D is a rich, interactive 3D environment based on human-designed indoor scenes sourced from SUNCG~\cite{song2017semantic}.

\smallskip
\noindent {\bf Modular Models.}
Neural module networks were originally introduced for visual question answering ~\cite{andreas2016neural}. 
These networks decompose a question into several components and dynamically assemble a network to compute the answer, dealing with variable compositional linguistic structures.
Since their introduction, modular networks have been applied to several other tasks: visual reasoning~\cite{hu2017learning, johnson2017inferring},  relationship modeling~\cite{ronghang16relationship}, embodied question answering~\cite{nmc}, multitask reinforcement learning~\cite{andreas2016modular}, language grounding on images~\cite{yu2018mattnet} and video understanding~\cite{feitemporal}. 
Inspired by~\cite{das2018neural, johnson2017inferring}, we cast EQA as a partially observable version of CLEVR and extend the modular idea to this task, which we believe requires an increasingly modular model design to address visual reasoning within a 3D environment. 


\vspace{-.1cm}
\section{Multi-Target EQA Dataset}
\label{sec:dataset}
\vspace{-.1cm}

\begin{table*}[t]
\footnotesize
\centering
\renewcommand{\arraystretch}{1.2}
\begin{tabular}{@{}l p{5.0cm} l@{}}
\toprule
& Question Type & Template\\
\midrule
\multirow{4}{*}{\rotatebox[origin=c]{90}{EQA-v1}} &
\textsf{location} & \myquote{What room is the $<$OBJ$>$ located in?}\\
&\textsf{color} & \myquote{What color is the $<$OBJ$>$?} \\
&\textsf{color\underline{ }room} & \myquote{What color is the $<$OBJ$>$ in the $<$ROOM$>$?} \\
&\textsf{preposition} & \myquote{What is $<$on/above/below/next-to$>$ the $<$OBJ$>$ in the $<$ROOM$>$?} \\
\midrule
\multirow{6}{*}{\rotatebox[origin=c]{90}{MT-EQA}} &
\textsf{object\_color\_compare\_inroom} & \myquote{Does $<$OBJ1$>$ share same color as $<$OBJ2$>$ in $<$ROOM$>$?} \\
&\textsf{object\_color\_compare\_xroom} & \myquote{Does $<$OBJ1$>$ in $<$ROOM1$>$ share same color as $<$OBJ2$>$ in $<$ROOM2$>$?} \\
&\textsf{object\_size\_compare\_inroom} & \myquote{Is $<$OBJ1$>$ bigger/smaller than $<$OBJ2$>$ in $<$ROOM$>$?}  \\
&\textsf{object\_size\_compare\_xroom} & \myquote{Is $<$OBJ1$>$ in $<$ROOM1$>$ bigger/smaller than $<$OBJ2$>$ in $<$ROOM2$>$?} \\
&\textsf{object\_dist\_compare} & \myquote{Is $<$OBJ1$>$ closer than/farther from $<$OBJ2$>$ than $<$OBJ3$>$ in $<$ROOM$>$?} \\
&\textsf{room\_size\_compare} & \myquote{Is $<$ROOM1$>$ bigger/smaller than $<$ROOM2$>$ in the house?} \\
\bottomrule
\end{tabular}
\vspace{0.05cm}
\caption{Question types and the associated templates used in EQA-v1 and MT-EQA. 
}
\label{table:question_types}
\end{table*}


We now describe our proposed Multi-Target Embodied Question Answering (MT-EQA) task and associated dataset, contrasting it against EQA-v1.
In v1~\cite{das2018embodied}, the authors select 750 (out of about 45,000) environments for the EQA task.
Four types of questions are proposed, each questioning a property (color, location, preposition) of a single target (room, object), as shown at the top of Table.~\ref{table:question_types}.
Our proposed MT-EQA task generalizes EQA-v1 and involves comparisons of various attributes (color, size, distance) between multiple targets, shown at the bottom of Table.~\ref{table:question_types}. Next, we describe in detail the generation process, as well as useful statistics of MT-EQA.

\begin{table*}[t]
\footnotesize
\centering
\begin{tabular}{@{}p{3.3cm}  l @{}}
\toprule
Question Type & Functional Form \\
\midrule
\textsf{object\underline{ }color\underline{ }compare} & \textsf{select(rooms) $\rightarrow$ unique(rooms) $\rightarrow$ select(objects) $\rightarrow$ unique(objects) $\rightarrow$ pair(objects) $\rightarrow$ query(color\underline{ }compare)} \\
\textsf{object\underline{ }size\underline{ }compare} & \textsf{select(rooms) $\rightarrow$ unique(rooms) $\rightarrow$ select(objects) $\rightarrow$ unique(objects) $\rightarrow$ pair(objects) $\rightarrow$query(size\underline{ }compare)} \\
\textsf{object\underline{ }dist\underline{ }compare} & \textsf{select(rooms) $\rightarrow$ unique(rooms) $\rightarrow$ select(objects) $\rightarrow$ unique(objects) $\rightarrow$ triplet(objects) $\rightarrow$query(dist\underline{ }compare)} \\
\textsf{room\underline{ }size\underline{ }compare} & \textsf{select(rooms) $\rightarrow$ unique(rooms) $\rightarrow$ pair(rooms) $\rightarrow$ query(size\underline{ }compare)} \\
\bottomrule
\end{tabular}
\vspace{0.05cm}
\caption{Functional forms of all question types in the MT-EQA dataset. Note that for each object color/size comparison question type, there exists two modes: \textsf{\small inroom} and \textsf{\small xroom}, depending on whether the two objects are in the same room or not. For example, \textsf{\small object\_color\_compare\_xroom} compares the color of two objects in two different rooms.}
\label{table:question_programs}
\end{table*}

\subsection{Multi-Target EQA Generation}
\vspace{-.1cm}
We generate question-answer pairs using the annotations available on SUNCG. We use the same number of rooms and objects as EQA-v1 (see Figure 2 in~\cite{das2018embodied}). Each question in MT-EQA is represented as a series of functional programs, which can be executed on the environment to yield a ground-truth answer.
The functional programs consist of some elementary operations, e.g., \textsf{\small select()}, \textsf{\small unique()}, \textsf{\small object\_color\_pair()}, \textsf{\small query()}, \etc, that operate on the room and object annotations.

Each question type is associated with a question template and a sequence of operations. For example, consider the question type in MT-EQA \textsf{\small object\_color\_compare}, whose template is \emph{``Does $<$OBJ1$>$ share same color as $<$OBJ2$>$ in $<$ROOM$>$?"}.
Its sequence of elementary operations is:

\noindent \textsf{\small select(rooms) $\rightarrow$ unique(rooms) $\rightarrow$ select(objects) $\rightarrow$ unique(objects) $\rightarrow$ pair(objects) $\rightarrow$ query(color\underline{ }compare)}.

The first function, \textsf{\small select(rooms)}, returns all rooms in the environment.
The second function, \textsf{\small unique(rooms)}, selects a single unique room from the list to avoid ambiguity.
Similarly, the third function, \textsf{\small select(objects)}, and fourth function, \textsf{\small unique(objects)}, return unique objects in the selected room.
The fifth function, \textsf{\small pair(objects)}, pairs the objects.
The final function, \textsf{\small query(color\underline{ }compare)}, compares their colors.

\begin{table}[t]
\centering
\resizebox{0.85\columnwidth}{!}{%
\begin{tabular}{@{}p{2cm}  c c  c  c  c@{}}
\toprule
& random & q-LSTM & q-NN & q-BoW & ``no" \\
\cmidrule{2-6}
Test Acc.~($\%$) & 49.44 & 48.24 & 53.74 & 49.22 & 53.28 \\
\bottomrule
\end{tabular}
}
\vspace{0.05cm}
\caption{EQA (test) accuracy using questions and priors.}
\label{tab:q_only}
\end{table}

We design 6 types of questions comparing different attributes between objects (inside same room/across different rooms), distance comparison, and room size comparison. All question types and templates are shown in Table~\ref{table:question_programs}.

In some cases, a question instantiation returned from the corresponding program, as shown above, might not be executable, as rooms might be disconnected or not reachable. 
To check if a question is feasible, we execute the corresponding \texttt{nav\_room()} and \texttt{nav\_object()} programs and compute shortest paths connecting the targets in the question.
If there is no path\footnote{This is a result of noisy annotations in SUNCG and inaccurate occupancy maps due to the axis-aligned assumption returned by House3D.}, it means the agent would not be able to look at all targets starting from its given spawn location. 
We filter out such impossible questions.

\begin{figure}[t]
\centering
\includegraphics[width=0.44\textwidth]{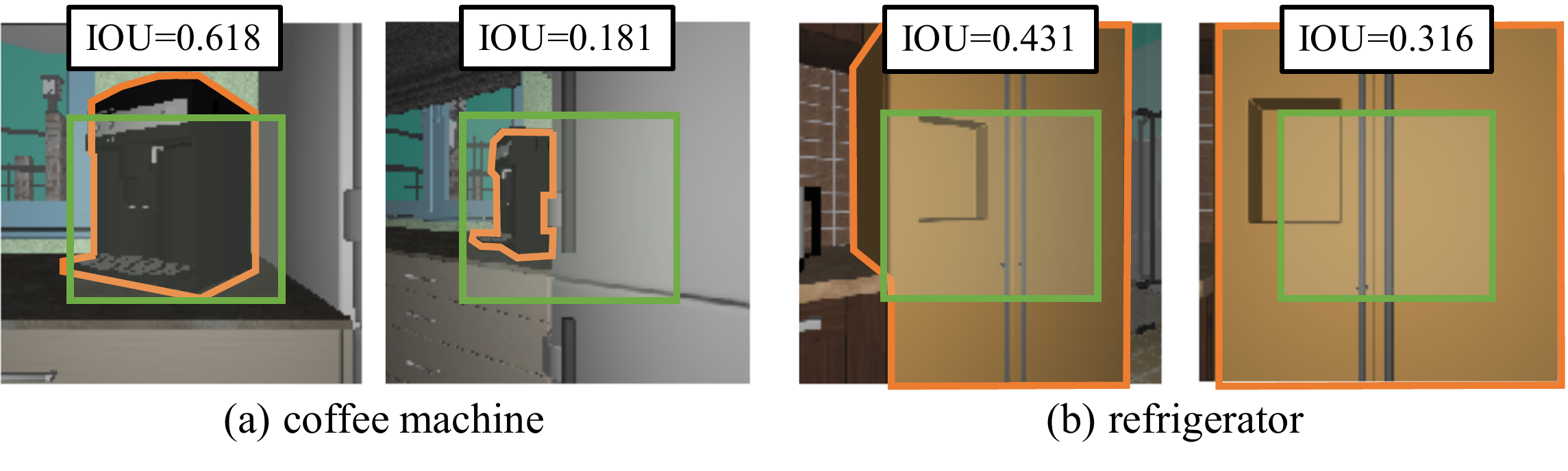}
\caption{IOU between the target's mask and the centered rectangle mask. Higher IOU is achieved when the target has larger portion in the center of the view.}
\label{fig:ious}
\end{figure}

For computing the shortest path connecting the targets, we need to find the position $(x, y, z, yaw)$ that best views each target. 
In order to do so, we first sample 100 positions near the target.
For each position, we pick the yaw angle that looks at the target with the highest Intersection-Over-Union (IOU), computed using the target's mask\footnote{House3D returns the the ground-truth semantic segmentation for each first-person view.} and a centered rectangular mask.
Fig.~\ref{fig:ious} shows 4 IOU scores of \emph{coffee machine} and \emph{refrigerator} from different positions. 
We sort the 100 positions and pick the one with highest IOU as the best-view position of the target, which is used to connect the shortest-path.
For each object, its highest IOU value $\textrm{IOU}_{best}$ is recorded for evaluation purposes (as a reference of the target's best-view).

\begin{figure}[t]
\centering
\includegraphics[width=0.48\textwidth]{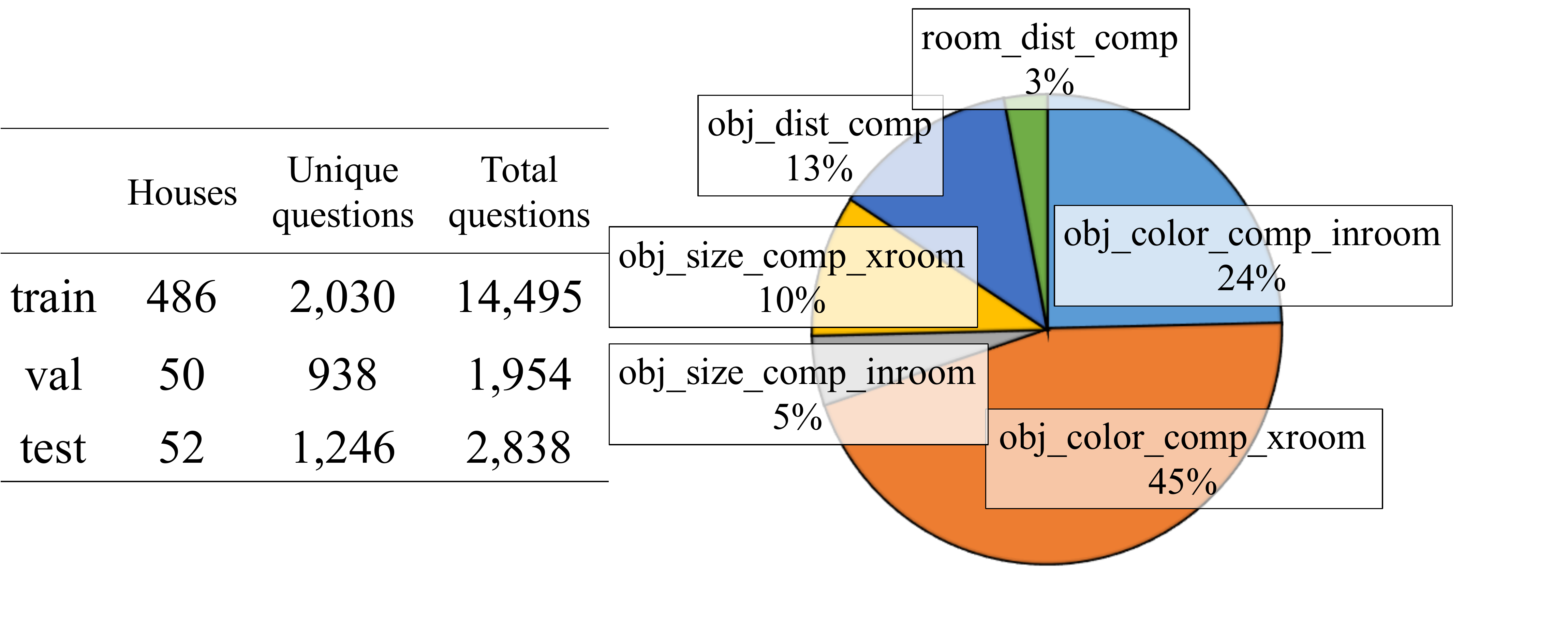}
\caption{Overview of MT-EQA dataset including split statistics and question type distribution.}
\label{fig:pie}
\end{figure}

To minimize the bias in MT-EQA, we perform entropy-filtering, similar to~\cite{das2018embodied}.
Specifically for each unique question, we compute its answer distribution across the whole dataset. 
We exclude questions whose normalized answer distribution entropy is below 0.9\footnote{Rather than 0.5 in~\cite{das2018embodied}, we set the normalized entropy threshold as 0.9 (maximum is 1) since all of our questions have binary answers.}.
This prevents the agent from memorizing easy question-answer pairs without looking at the environment.
For example, the answer to \myquote{is the bed in the living room bigger than the cup in the kitchen?} is always \emph{Yes}. Such questions are excluded from our dataset.
After the two filtering stages, the MT-EQA questions are both balanced and feasible.

\begin{figure*}[t]
\centering
\includegraphics[width=0.90\textwidth]{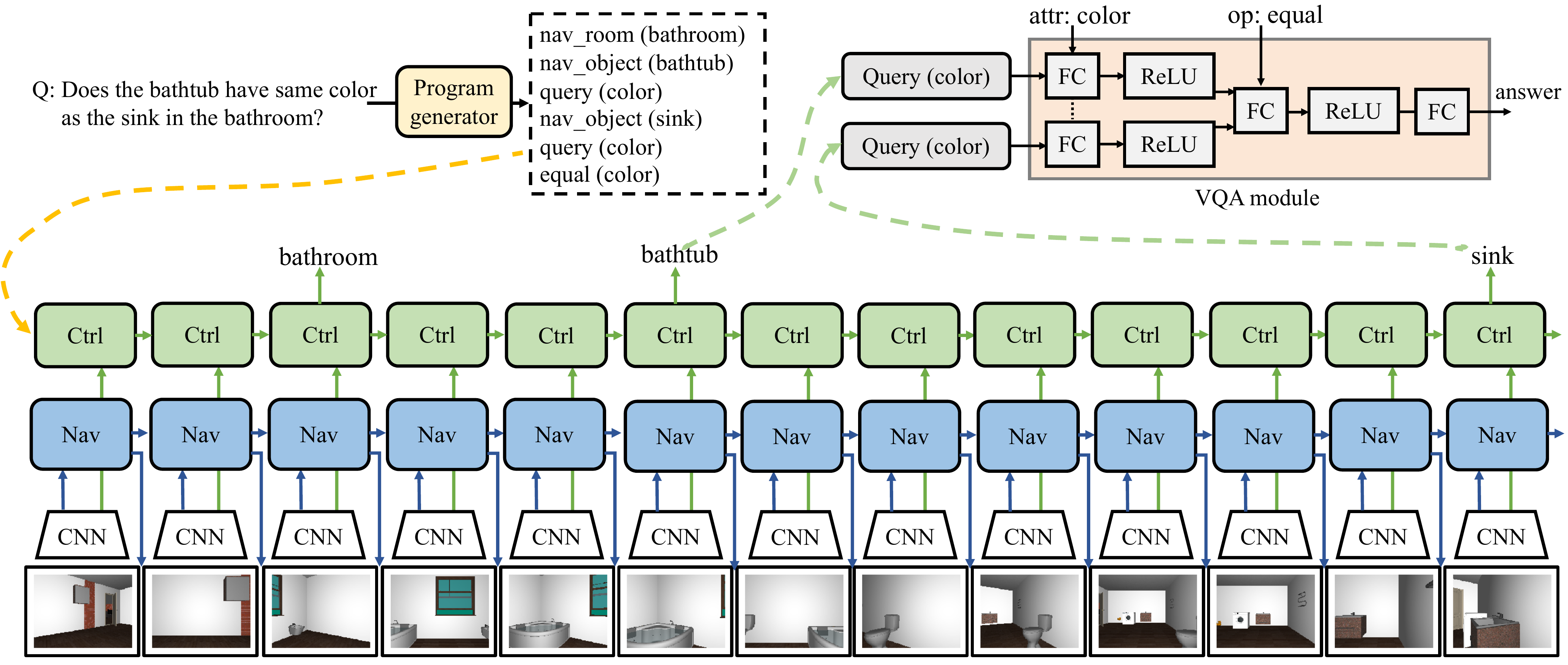}
\caption{Model architecture: our model is composed of a program generator, a navigator, a controller, and a VQA module.}
\label{fig:model}
\end{figure*}

In addition, we check if MT-EQA is easily addressed by question-only or prior-only baselines. For this, we evaluate four question-based models: (a) an LSTM-based question-to-answer model, (b) a nearest neighbor (NN) baseline that finds the NN question from the training set and uses its most frequent answer as the prediction, (c) a bag-of-words (BoW) model that encodes a question followed by a learned linear classifier to predict the answer and (d) a naive ``no" only answer model, since ``no" is the most frequent answer by a slight margin.
Table.~\ref{tab:q_only} shows the results. 
There exists very little bias on the ``yes/no" distribution (53.28\%), and all question-based models make close to random predictions.
In comparison, and as we empirically show in Sec.~\ref{sec:experiments}, our results are far better than these baselines, indicating the necessity to explore the environment in order to answer the question. 
Besides, the results also address the concern in~\cite{anand2018} where language-only models (BoW and NN) already form competitive baselines for EQA-v1. 
In MT-EQA, these baselines perform close to chance as a result of the balanced binary question-answer pairs in MT-EQA.

Overall, our MT-EQA dataset consists of 19,287 questions across 588 environments\footnote{The 588 environments are subset of EQA-v1's. Some environments are discarded due to entropy filtering and unavailable paths.}, referring to a total of 61 unique object types in 8 unique room types.
Fig.~\ref{fig:pie} shows the question type distribution.
Approximately 32 questions are asked for each house on average, 209 at most and 1 at fewest.
There are relatively fewer \textsf{\small object\_size\_compare} and \textsf{\small room\_size\_compare} questions as many frequently occurring comparisons are too easy to guess without exploring the environment and thus fail the entropy filtering. We will release the MT-EQA dataset and the generation pipeline.

\vspace{-.1cm}
\section{Model}
\label{sec:model}
\vspace{-.1cm}

\begin{table}[t]
\small
\centering
\resizebox{0.82\columnwidth}{!}{%
\begin{tabular}{@{}l l@{}}
\toprule
1) \texttt{nav\_object(phrase)} & 2) \texttt{nav\_room(phrase)} \\
\multicolumn{2}{@{}l@{}}{3) \texttt{query(color / size / room\_size)}}\\
\multicolumn{2}{@{}l@{}}{4) \texttt{equal\_color()}} \\
\multicolumn{2}{@{}l@{}}{5) \texttt{object\_size\_compare(bigger / smaller)}} \\
\multicolumn{2}{@{}l@{}}{6) \texttt{object\_dist\_compare(farther / closer)}} \\
\multicolumn{2}{@{}l@{}}{7) \texttt{room\_size\_compare(bigger / smaller)}} \\
\bottomrule
\end{tabular}
}
\vspace{0.07cm}
\caption{MT-EQA executable programs.}
\label{tab:pgs}
\end{table}

Our model is composed of 4 modules: the question-to-program generator, the navigator, the controller, and the VQA module.
We describe these modules in detail.

\subsection{Program Generator}
\vspace{-.1cm}
The program generator takes the question as input and generates sequential programs for execution.
We define 7 types of executable programs for the MT-EQA task in Table.~\ref{tab:pgs}.
For example, \myquote{Is the bathtub the same color as the sink in the bathroom?} is decomposed into a series of sequential sub-programs:
\noindent \texttt{nav\_room(bathroom)} $\rightarrow$ \texttt{nav\_object(bathtub)} $\rightarrow$ \texttt{query\_color()} $\rightarrow$ \texttt{nav\_object(sink)} $\rightarrow$ \texttt{query\_color()} $\rightarrow$ \texttt{equal\_color()}.
Similar to CLEVR~\cite{johnson2017clevr}, the question programs are automatically generated in a templated manner (Table.~\ref{table:question_programs}), making sub-component decomposition (converting questions back to programs) simple (Table.~\ref{tab:pgs}).
We use template-based rules by selecting and filling in the arguments in Table.~\ref{tab:pgs} to generate the programs (which is always accurate).
While a neural model could also be applied, a learned program generator is not the focus of our work. 


\vspace{-.1cm}
\subsection{Navigator}
\begin{figure}[t]
    \centering
    \begin{subfigure}[t]{0.23\textwidth}
        \centering
        \includegraphics[height=1.25in]{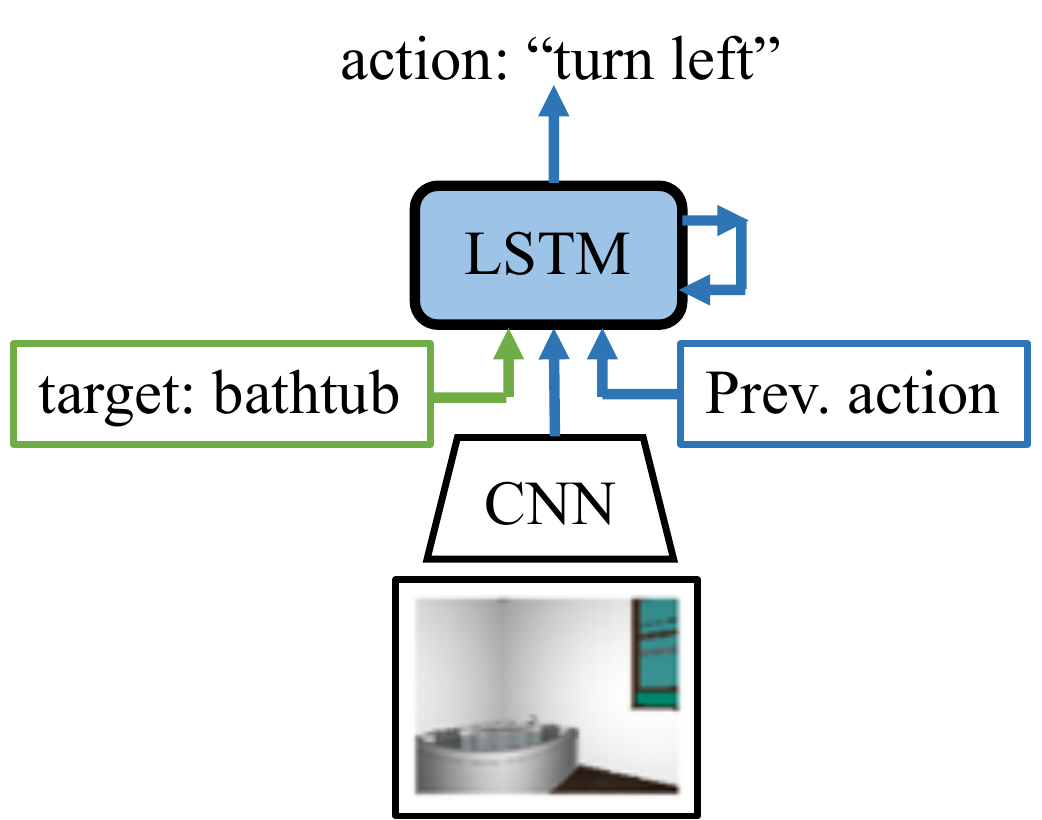}
        \caption{Navigator}
    \end{subfigure}%
    \begin{subfigure}[t]{0.23\textwidth}
        \centering
        \includegraphics[height=1.25in]{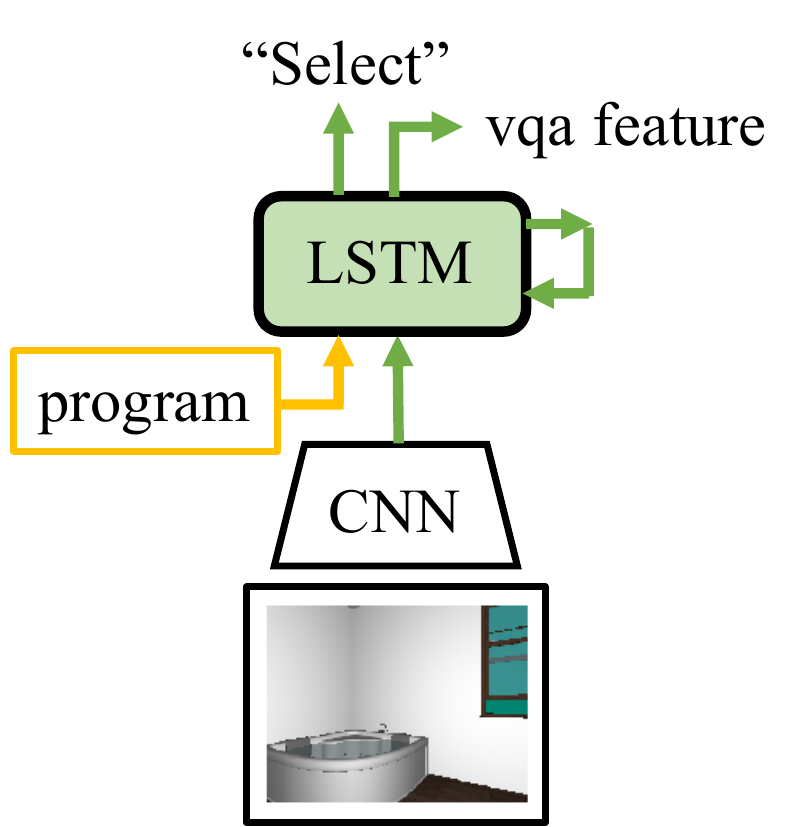}
        \caption{Controller}
    \end{subfigure}
    \vspace{0.06cm}
    \caption{Navigator and Controller.}
    \label{fig:nav_ctrl}
\end{figure}

\vspace{-.1cm}
The navigator executes the \texttt{nav\_room()} and \texttt{nav\_object()} programs.
As shown in Fig.~\ref{fig:nav_ctrl}(a), we use an LSTM as our core component.
At each time step, the LSTM takes as inputs the current egocentric (first-person view) image, an encoding of the target phrase (\eg ``bathtub" if the program is \texttt{nav\_object(bathtub)}), and the previous action, in order to predict the next action.

The navigator uses a CNN feature extractor that takes a 224x224 RGB image returned from the House3D renderer, and transforms it into a visual feature, which is then fed into the LSTM.
Similar to~\cite{das2018embodied}, the CNN is pre-trained under a multi-task framework consisting of three tasks: RGB-value reconstruction, semantic segmentation, and depth estimation. 
Thus, the extracted feature contains rich information about the scene's appearance, content, and geometry (objects, color, texture, shape, and depth).
In addition to the visual feature, the LSTM is presented with two additional inputs. 
The first is the target embedding, where we use the average embedding of GloVE vectors~\cite{pennington2014glove} over words describing the target. 
The second is previous action, which is in the form of a look-up from an action embedding matrix.

We want to note the different perceptual skills required for room and object navigation: 
Room navigation relies on understanding the overall scene and finding cross-room paths (entry/exit), while object navigation requires localizing the target object within a room and finding a path to reach it.
To capture the difference, we implement two separate navigation modules, \texttt{nav\_room()} and \texttt{nav\_object()} respectively. 
These two modules share same architecture but are trained separately for different targets. 

In MT-EQA, the action space for navigation consists of 3 action types: turning left (30 degrees), turning right (30 degrees), and moving forward. 
This is almost the same as EQA-v1~\cite{das2018embodied}, except we use larger turning angles -- as our navigation paths are much longer due to the multi-target setting. 
We find that this change reduces the number of actions required for navigation, leading to easier training.

\vspace{-.1cm}
\subsection{Controller}
\vspace{-.1cm}
The controller is the central module in our model, as it connects all of the other modules by: 1) creating a plan from the program generator, 2) collecting the necessary observations from the navigator, and 3) invoking the VQA module.

Fig.~\ref{fig:nav_ctrl} (b) shows the controller, whose key component is another LSTM. 
Consider the question \myquote{Does the bathtub have same color as the sink in the bathroom?} with part of its program as example -- \texttt{nav\_room(bathroom)} $\rightarrow$ \texttt{nav\_object(bathtub)}. 
The controller starts by calling the room navigator to look for ``bathroom''.
During navigation, the controller keeps track of the first-person views, looking for the target. 
Particularly, it extracts the features via CNN which are then fused with the target embedding as input to the LSTM. 
The controller predicts SELECT if the target is found, stopping the current navigator, in our example \texttt{nav\_room(bathroom)}, and starting execution of the next program, \texttt{nav\_object(bathtub)}.

Finally, after the object target ``bathtub" has been found, the next program -- \texttt{query\_color()}, is executed. 
The controller extracts attribute features from the first-person view containing the target.
In all, there are three attribute types in MT-EQA - object's color, object's size, and room's size. 
Again, we treat object and room differently in our model.
For object-specific attributes, we use the hidden state of the controller at the location where SELECT was predicted. 
This state should contain semantic information for the target, as it is where the controller is confident the target is located. 
For room-specific attributes, the controller collects a panorama by asking the navigator to rotate 360 degrees (by performing 12 turning-right actions) at the location where SELECT is predicted. 
The CNN features from this panorama view are concatenated as the representation. 

During program execution by the controller, the extracted cues for all the targets are stored, and in the end they are used by the VQA module to predict the final answer.

\vspace{-.1cm}
\subsection{VQA Module}
\vspace{-.1cm}
The final task requires comparative reasoning, \eg, \texttt{\small{object\_size\_compare(bigger)}}, \texttt{\small{equal\_color()}}, \etc.
When the controller has gathered all of the targets for comparison, it invokes the VQA module.
As shown in top-right of Fig.~\ref{fig:model}, the VQA module embeds the stored features of multiple targets into the question-attribute space, using a FC layer followed by ReLU.
The transformed features are then concatenated and fed into another FC+ReLU which is conditioned on the comparison operator (equal, bigger than, smaller than, \etc).
The output is a binary prediction (yes/no) for that attribute comparison.
We call it compositional VQA (cVQA).
The cVQA module in Fig.~\ref{fig:model} depicts a two-input comparison as an example, but our cVQA module also extends to three inputs, for questions like \myquote{Is the refrigerator closer to the coffee machine than the microwave?}.

\vspace{-.1cm}
\subsection{Training}
\vspace{-.1cm}
Training follows a two-stage approach: First, the full model is trained using Imitation Learning (IL); Second, the navigator is further fine-tuned with Reinforcement Learning (RL) using policy gradients.

First, we jointly train our full model using imitation learning. For imitation learning, we treat the shortest paths and the key positions containing the targets as our ground-truth labels for navigation and for the controller's SELECT classifier, respectively. 
The objective function consists of a navigation objective and a controller objective at every time step $t$, and a VQA objective at the final step.
For the i-th question, let $P^{nav}_{i,t,a}$ be action $a$'s probability at time $t$, $P^{sel}_{i,t}$ be the controller's SELECT probability at time $t$, and $P^{vqa}_i$ be the answer probability from VQA, then we minimize the combined loss:

\begin{equation}\nonumber
\begin{split}
L &= L_{nav} + \alpha L_{ctrl} + \beta L_{vqa} \\
&= \resizebox{0.50\hsize}{!}{ $\underbrace{-\sum_i \sum_t \sum_a y^n_{i,t,a} \log P^{nav}_{i,t,a}}_{\text{Cross-entropy on navigator action} }  $}\\
& \resizebox{0.96\hsize}{!}{$ \quad-\alpha \underbrace{ \sum_i \sum_t (y^c_{i,t} \log P^{sel}_{i,t} + (1-y^c_{i,t}) \log( 1 - P^{sel}_{i,t}) ) }_{\text{Binary cross-entropy on controller's SELECT}}  $} \\
& \resizebox{0.86\hsize}{!}{$ \quad-\beta \underbrace{ \sum_{i} (y^v_i \log P^{vqa}_i + (1-y^v_i) \log( 1 - P^{vqa}_i) )}_{\text{Binary cross-entropy on VQA's answer}} $}.
\end{split}
\label{eq:objective}
\end{equation}
\vspace{-0.3cm}

Subsequently, we use RL to fine-tune the room and object navigators.

We provide two types of reward signals to the navigators. 
The first is a dense reward, corresponding to the agent's progress toward the goal (positive if moving closer to the target and negative if moving away). 
This reward is measured by the distance change in the 2D bird-view distance space, clipped to lie within $[-1.0, 1.0]$.
The second is a sparse reward that quantifies whether the agent is looking at the target object when the episode is terminated.
For object targets, we compute $\textrm{IOU}_T$ between the target's mask and the centered rectangle mask at termination.
We use the best IOU score of the target $\textrm{IOU}_{best}$ as reference and compute the ratio $\frac{\textrm{IOU}_T}{\textrm{IOU}_{best}}$.
If the ratio is greater than 0.5, we set the reward to 1.0 otherwise -1.0.
For room targets, we assign reward 0.2 to the agent if it is inside the target room at termination, otherwise -0.2.


\vspace{-.1cm}
\section{Experiments}\label{sec:experiments}
\vspace{-.1cm}

\begin{table*}[t]
\footnotesize
\centering
\resizebox{1.9\columnwidth}{!}{%
\begin{tabular}{@{} c  l  c c c c c   c c  c c c c c@{}}
\toprule
\multicolumn{2}{c}{} & \multicolumn{5}{c}{Object Navigation} & \multicolumn{2}{c}{Room Navigation} & \multicolumn{5}{c}{EQA}\\
\cmidrule(lr){3-7} \cmidrule(lr){8-9} \cmidrule(l){10-14}
\multicolumn{2}{c}{} & $d_T$ & $d_{\Delta}$ & $h_T$ & $IOU^r_T$ & $\%stop_{o}$ & $\%r_T$ & $\%stop_{r}$ & $ep\_len$ & $\%_{easy}$ & $\%_{medium}$ & $\%_{hard}$ & $\%_{overall}$ \\
\midrule
1 & \mbox{Nav+cVQA}       & 5.41 & -0.64 & 0.19 & 0.15 & 36 & 34 & 60 & 153.13 & 58.42 & 53.29 & 51.46 & 53.24 \\
2 & \mbox{Nav(RL)+cVQA}  & 3.80 & 0.10 & 0.33 & \textbf{0.30} & 46 & 40 & 62 & 144.80 & 67.57 & 55.91 & 53.28 & 57.40 \\
3 & \mbox{Nav+Ctrl+cVQA}  & 5.25 & -0.56 & 0.20 & 0.18 & 36 & 37 & 70 & 145.20 & 59.73 & 53.48 & 49.04 & 54.44 \\
4 & \mbox{Nav(RL)+Ctrl+cVQA} & \textbf{3.60} & \textbf{0.16} & \textbf{0.33} & 0.29 & \textbf{48} & \textbf{43} & \textbf{72} & \textbf{127.71} & \textbf{72.22} & \textbf{59.97} & \textbf{54.92} & \textbf{61.45} \\
\bottomrule
\vspace{-.2cm}
\end{tabular}
}
\caption{Quantitative evaluation of object/room navigation and EQA accuracy for different approaches.}
\label{table:EQA_whole}
\end{table*}

\begin{table*}[t]
\footnotesize
\centering
\resizebox{1.9\columnwidth}{!}{%
\begin{tabular}{@{} c  l   c c   c c   c   c  c@{}}
\toprule
\multicolumn{2}{c}{}  & \multicolumn{2}{c}{$\textsf{object\_color\_compare}$} & \multicolumn{2}{c}{$\textsf{object\_size\_compare}$} & $\textsf{object\_dist\_compare}$ & $\textsf{room\_size\_compare}$ & \multirow{2}{*}{$\%_{overall}$} \\
\cmidrule(lr){3-8} 
\multicolumn{2}{c}{} & \mbox{inroom} & \mbox{xroom} & \mbox{inroom} & \mbox{xroom} & \mbox{inroom} & \mbox{xroom}\\
\midrule
1 & \mbox{Nav+cVQA}    & 64.15 & 52.47 & 57.85 & 55.68 & 49.38 & 48.37 & 53.24  \\
2 & \mbox{Nav(RL)+cVQA}    & 71.24 & 53.92 & 74.38 & 60.81 & 51.23 & 46.66 & 57.40 \\
3 & \mbox{Nav+Ctrl+cVQA} & 66.41 & 52.65 & 57.85 & 53.48 & 49.38 & 48.37 & 54.44  \\
4 & \mbox{Nav(RL)+Ctrl+cVQA} & \textbf{72.68} & \textbf{58.19} & \textbf{76.86} & \textbf{63.37} & \textbf{54.94} & \textbf{55.57} & \textbf{61.45}  \\
\bottomrule
\vspace{-.2cm}
\end{tabular}
}
\caption{EQA accuracy on each question type for different approaches.}
\label{table:EQA_each_type}
\end{table*}

\begin{table*}[h!]
\footnotesize
\centering
\resizebox{1.9\columnwidth}{!}{%
\begin{tabular}{@{} c  l   c c   c c   c   c   c@{}}
\toprule
\multicolumn{2}{c}{}  & \multicolumn{2}{c}{$\textsf{object\_color\_compare}$} & \multicolumn{2}{c}{$\textsf{object\_size\_compare}$} & $\textsf{object\_dist\_compare}$ & $\textsf{room\_size\_compare}$ & \multirow{2}{*}{$\%_{overall}$} \\
\cmidrule(lr){3-8}
\multicolumn{2}{c}{} & \mbox{inroom} & \mbox{xroom} & \mbox{inroom} & \mbox{xroom} & \mbox{inroom} & \mbox{xroom}\\
\midrule
1 & \mbox{[BestView] + attn-VQA (cnn)} & 71.16 & 59.56 & 65.29 & 65.93 & 58.64 & 49.74 & 60.50 \\
2 & \mbox{[BestView] + cVQA (cnn)} & 82.92 & 72.70 & 80.99 & 83.88 & \textbf{69.75} & 64.32 & 74.14 \\
3 & \mbox{[ShortestPath+BestView] + Ctrl + cVQA} & \textbf{90.70} & \textbf{85.4}9 & \textbf{82.64} & \textbf{88.64} & 68.52 & \textbf{71.87} & \textbf{82.88} \\
4 & \mbox{[ShortestPath] + seq-VQA} & 53.32 & 54.44 & 51.24 & 50.55 & 47.53 & 49.74 & 52.36 \\
5 & \mbox{[ShortestPath] + Ctrl + cVQA} & 76.09 & 69.11 & 75.21 & 79.49 & 64.20 & 61.23 & 69.77 \\
\bottomrule
\vspace{-.2cm}
\end{tabular}
}
\caption{EQA accuracy of different approaches on each question type in oracle setting (given shortest path or best-view images).}
\label{table:EQA_oracle}
\end{table*}

In this section we describe our experimental results.
Since MT-EQA is a complex task and our model is modular, we will show both the final results (QA accuracy) and the intermediate performance (for navigation). Specifically, we first describe our evaluation setup and metrics for MT-EQA. Then, we report the comparison of our model against several strong baselines. And finally, we analyze variants of our model and provide ablation results.

\vspace{-.1cm}
\subsection{Evaluation Setup and Metrics}
\vspace{-.1cm}
\noindent\textbf{Spawn Location}. MT-EQA questions involve multiple targets (rooms/objects) to be found.
To prevent the agent from learning biases due to spawn location, we randomly select one of the mentioned targets as reference and spawn our agent 10 actions (typically 1.9 meters) away.
\\
\noindent\textbf{EQA Accuracy}. We compute overall accuracy as well as accuracy for each of the 6 types of questions in our dataset.
In addition, we also categorize question difficulty level into easy, medium, and hard by binning the ground-truth action length.
Easy questions are those with fewer than 25 action steps along the shortest path, medium are those with 25-70 actions, and hard are those with more than 70 actions.
We report accuracy for each difficulty, $\%_{easy}$, $\%_{medium}$, $\%_{hard}$, as well as overall, $\%_{overall}$, in Table~\ref{table:EQA_whole}.
\\
\noindent\textbf{Navigation Accuracy}.
We also measure the navigation accuracy for both objects and rooms in MT-EQA.
As each question involves several targets, the order of them being navigated matters.
We consider the `ground truth' ordering of targets for navigation as the order in which they are mentioned in the question, e.g., given \emph{``Does the bathtub have same color as the sink?"}, the agent is trained and evaluated for visiting the ``bathtub" first and then the ``sink".

For each mentioned target object, we evaluate the agent's navigation performance by computing the distance to the target object at navigation termination, $d_T$, and change in distance to the target from initial spawned position to terminal position, $d_{\Delta}$. 
We also compute the stop ratio $\%stop_{o}$ as in EQA-v1~\cite{das2018embodied}.
Additionally, we propose two new metrics based on the IOU of the target object at its termination.
When the navigation is done, we compute the IOU of the target w.r.t a centered rectangular box (see Fig.~\ref{fig:ious} as example).
The first metric is mean IOU ratio 
$\textrm{IOU}_T^r=\frac{1}{N}\sum_i\frac{\textrm{IOU}_T(o_i)}{\textrm{IOU}_{best}(o_i)})$ where $\textrm{IOU}_{best}(o_i)$ is the highest IOU score for object $o_i$.
The second is hit accuracy $h_T$ -- we compute the percentage of the ratio $\textrm{IOU}_T(o_i)/\textrm{IOU}_{best}(o_i)$ greater than 0.5, i.e., $h_T = \frac{1}{N}\sum_i ||\frac{\textrm{IOU}_T(o_i)}{\textrm{IOU}_{best}(o_i)}>0.5||$.
Both metrics measure to what extent the agent is looking at the target at termination.

For each mentioned target room, we evaluate the agent's navigation by recording the percentage of agents terminating inside the target room $\%r_T$ and the stop ratio $\%stop_{r}$.

For all the above metrics except for $d_T$, larger is better.
Additionally, we report the overall number of action steps (episode length) executed for each question, i.e., $ep\_len$.

\vspace{-.1cm}
\subsection{EQA Results}
\vspace{-.1cm}
Nav+Ctrl+cVQA is our full model, which is composed of a program generator, a navigator, a controller and a comparative VQA module.
Another variant of our model, the REINFORCE fine-tuned model is denoted as Nav(RL)+Ctrl+cVQA.
We also train a simplified version of our full model, Nav+cVQA. which does not use a controller.
For this model, we let the navigator predict termination whenever a target is detected, then feed its hidden states to the VQA model.
The training details are similar to our full model for both IL and RL.
We show  comparisons of both navigation and EQA accuracy in Table.~\ref{table:EQA_whole}.
\\
\noindent \textbf{RL helps both navigation and EQA accuracies}. 
Both object and room navigation performance are improved after RL finetuning.
We notice without finetuning $d_{\Delta}$ for both models (Row 1 \&3 ) are negative, which means the agent has moved farther away from the target during navigation.
After RL finetuning, $d_{\Delta}$ jumps from $-0.56$ to $0.16$ (Row 3 \& 4).
The hit accuracy also improves from 20\% to 33\%, indicating that the RL-finetuned agent is more likely to find the target mentioned in the question.
Episode lengths from the stronger navigators are shorter, indicating that better navigators find their target more quickly.
And, higher EQA accuracy is also achieved with the help of RL finetuning (from 54.44\% to 61.45\%).
After breaking down the EQA into different types, we observe the same trend in Table.~\ref{table:EQA_each_type} -- our full model with RL far outperforms the others.

\noindent \textbf{Controller is important.}
Comparing our full model (Row 4) to the one without a controller (Row 2), we notice that the former outperforms the latter across almost all the metrics.
One possible reason is that the VQA task and navigation task are quite different, such that the features (hidden state) from the navigator cannot help improve the VQA module. 
On the contrary, our controller decouples the two tasks, letting the navigator and VQA module focus on their own roles.

\noindent \textbf{Questions with shorter ground-truth path are easier.}
We observe that our agent is far better at dealing with easy questions than hard ones (72.22\% over 54.92\% in Table.~\ref{table:EQA_whole} Row 4).
One reason is that the targets mentioned in the easy questions, e.g., sink and toilet in \myquote{Does the sink have same color as the toilet in the bathroom?}, are typically closer to each other, thus are relatively easier to be explored, whereas questions like \myquote{Is the kitchen bigger than the garage?} requires a very long trajectory and the risk of missing one (kitchen or garage) is increased.
The same observation is found in Table.~\ref{table:EQA_each_type}, where we get higher accuracy for ``in-room'' questions than ``cross-room'' ones.

\vspace{-.1cm}
\subsection{Oracle Comparisons}
\vspace{-.1cm}
To better understand each module of our model, we run ablation studies.
Table.~\ref{table:EQA_oracle} shows EQA accuracy of different approaches given the shortest paths or best-view frames.

\noindent \textbf{Our VQA module helps.}
We first compare the performance of our VQA module against an attention-based VQA.
Given the best view of each target, we can directly feed the features from those images to the VQA module, using the CNN features instead of hidden states from controller side.
The attention-based VQA architecture is similar to~\cite{das2018embodied}, which uses an LSTM to encode questions and then uses its representation to pool image features with attention.
Comparing the two methods in Table.~\ref{table:EQA_oracle}, Row 1 \& 2, our VQA module achieves 13.64\% higher accuracy.
The benefit mainly comes from the decomposition of attribute representation and comparison in our VQA module.

\noindent \textbf{Controller's features help.}
We compare the controller's features to raw CNN features for VQA.
When given both shortest path and best-view position, we run our full model with these annotations and feed the hidden states from the controller's LSTM to our VQA model.
As shown in Table.~\ref{table:EQA_oracle}, Row 2 \& 3, the controller's features are far better than raw CNN features, especially for \textsf{\small object\_color\_compare} and \textsf{\small object\_size\_compare} question types.

\noindent \textbf{Controller's SELECT matters.} 
Our controller predicts SELECT and extracts the features at that moment.
One possible question is how important is this moment selection.
To demonstrate its advantage, we trained another VQA module which uses a LSTM to encode the whole sequence of frames along the shortest path and uses its final hidden state to predict the answer, denoted as seq-VQA.
The hypothesis is that the final hidden state might be able to encode all relevant information, as the LSTM has gone through the whole sequence of frames.
Table.~\ref{table:EQA_oracle}, Row 4, shows its results, which is nearly random.
On the contrary, when controller is used to SELECT frames in Row 5, the results are far better.
However, there is still much space for improvement.
Comparing Table.~\ref{table:EQA_oracle}, Row 3 \& 5, the overall accuracy drops 13\% when using features from the predicted SELECT instead of oracle moments, and 20\% when using additional navigators (comparing Table.~\ref{table:EQA_oracle}, Row 3, \& Table.~\ref{table:EQA_each_type}, Row 4), indicating the necessity of both accurate SELECT and navigation.


\vspace{-.1cm}
\section{Conclusion}
\vspace{-.1cm}
We proposed MT-EQA, extending the original EQA questions from a limited single-target setting to a more challenging multi-target setting, which requires the agent to perform comparative reasoning before answering questions. 
We collected a MT-EQA dataset as a test benchmark for the task, and validated its usefulness with simple baselines from just text or prior.
We also proposed a new EQA model consisting of four modular components: a program generator, a navigator, a controller, and VQA module for MT-EQA. 
We experimentally demonstrated that our model significantly outperforms baselines on both question answering and navigation, and conducted detailed ablative analysis for each component in both the embodied and oracle settings.

\smallskip
\noindent
{\bf Acknowledgements:}
We thank Abhishek Das, Devi Parikh and Marcus Rohrbach for helpful discussions.
This work is supported by NSF Awards \#1633295, 1562098, 1405822, and Facebook. 

{\small
\bibliographystyle{ieee}
\bibliography{egbib}
}

\end{document}